\documentclass[sigconf]{acmart}

\usepackage{booktabs} 
\usepackage{wrapfig}
\usepackage{hyperref}
\hypersetup{colorlinks=true}
\usepackage{subcaption}
\usepackage[labelfont=bf,textfont=bf]{caption}
\usepackage{graphicx}

\copyrightyear{2018} 
\acmYear{2018} 
\setcopyright{acmcopyright}
\acmConference[PETRA '18]{The 11th PErvasive Technologies Related to Assistive Environments Conference}{June 26--29, 2018}{Corfu, Greece}
\acmBooktitle{PETRA '18: The 11th PErvasive Technologies Related to Assistive Environments Conference, June 26--29, 2018, Corfu, Greece}
\acmPrice{15.00}
\acmDOI{10.1145/3197768.3201538}
\acmISBN{978-1-4503-6390-7/18/06}

\editor{Jennifer B. Sartor}
\editor{Theo D'Hondt}
\editor{Wolfgang De Meuter}

\begin{document}
\title{Towards Deep Learning based Hand Keypoints Detection for Rapid Sequential Movements from RGB Images}
\author{Srujana Gattupalli}
\affiliation{%
  \institution{Vision-Learning-Mining Lab,\\
  University of Texas at Arlington}
  \city{Arlington} 
  \state{Texas} 
}
\email{srujana.gattupalli@mav.uta.edu}

\author{Ashwin Ramesh Babu}
\affiliation{%
  \institution{Heracleia - Human Centered Computing Lab,\\
  University of Texas at Arlington}
  \city{Arlington} 
  \state{Texas} 
}
\email{ashwin.rameshbabu@mavs.uta.edu}

\author{James Robert Brady}
\affiliation{%
  \institution{Heracleia - Human Centered Computing Lab,\\
  University of Texas at Arlington}
  \city{Arlington} 
  \state{Texas} 
}
\email{james.brady2@mavs.uta.edu}

\author{Fillia Makedon}
\affiliation{%
  \institution{Heracleia - Human Centered Computing Lab,\\
  University of Texas at Arlington}
  \city{Arlington} 
  \state{Texas} 
}
\email{makedon@uta.edu}

\author{Vassilis Athitsos}
\affiliation{%
  \institution{Vision-Learning-Mining Lab,\\
  University of Texas at Arlington}
  \city{Arlington} 
  \state{Texas} 
}
\email{athitsos@uta.edu}
\renewcommand{\shortauthors}{Gattupalli et al.}

\begin{abstract}
Hand keypoints detection and pose estimation has numerous applications in computer vision, but it is still an unsolved problem in many aspects. An application of hand keypoints detection is in performing cognitive assessments of a subject by observing the performance of that subject in physical tasks involving rapid finger motion. As a part of this work, we introduce a novel hand keypoints benchmark dataset that consists of hand gestures recorded specifically for cognitive behavior monitoring. We explore the state of the art methods in hand keypoint detection and we provide quantitative evaluations for the performance of these methods on our dataset. In future, these results and our dataset can serve as a useful benchmark for hand keypoint recognition for rapid finger movements.
\end{abstract}

%
%

\ccsdesc[100]{Human-Centered Computing~Gestural input}
\ccsdesc[100]{Human-Centered Computing~Information Visualization}
\ccsdesc[100]{Computer Vision~Scene understanding}
\ccsdesc[100]{Computer Vision~Activity recognition and understanding}
\ccsdesc[100]{Computer Vision~Object Recognition}
\ccsdesc[100]{Computer Vision~Tracking}
\ccsdesc[100]{Machine Learning~Supervised learning by regression}
\ccsdesc[100]{Machine Learning~Transfer Learning}
\ccsdesc[100]{Machine Learning~Neural Networks}

\keywords{hand keypoints detection, deep learning, computer vision, convolutional neural networks, hand pose recognition, cognitive behavior assessment, gesture recognition}

\maketitle

\section{Introduction}
  
Hand pose estimation is an important computer vision topic  due to its wide range of applications in human-computer interaction, augmented/virtual reality, and gaming. Such applications often require hand segmentation, articulated hand pose estimation and tracking.  Recent methods in body pose estimation \cite{insafutdinov2016deepercut,openpose-simon2017hand} can be used to detect and segment hands using human body hand joint features. Articulated hand pose estimation from monocular RGB images is still a largely unsolved problem in several aspects. This is because human hand configurations are very diverse, with the pose of a human hand having over 20 Degrees of Freedom (DoF). Hands are smaller that the body, and thus they occupy a small part of the image when the full body is visible. In addition, hand keypoints are oftentimes occluded by other parts of the same hand, the other hand, or the rest of the body.

Deep learning based methods currently achieve state of the art performance for human body pose estimation. Estimating body pose is an articulated pose estimation problem similar to hand pose estimation. However, body pose estimation is easier, due to the body orientation being upright most time, and also due to occlusions being a less frequent and less severe problem for full body images compared to hand images. We investigate deep learning based methods for hand pose estimation that perform holistic articulated pose estimation. Pixel-wise pose estimation approaches could be slow for real-time applications and do not take advantage of important holistic hand features due to per pixel constraints.

In this work, we focus on RGB-based articulated hand pose estimation. We prefer this modality due to the availability and ease of deployment of regular color cameras compared to depth cameras. Our contribution aims towards partial hand pose estimation problem on single RGB image frames. The hand keypoints that we estimate are wrist, finger tips for thumb, index finger, middle finger, ring finger, little finger. We provide a novel RGB benchmark dataset for hand keypoints estimation and perform evaluations to provide quantitative evaluation for current state-of-the-art methods for this task. This dataset includes hand gestures and keypoint annotations for gestures pertaining to rhythmic hand movements. Our motivation is that tasks involving such movements can be used for cognitive assessments, in conjunction with tasks involving whole body motion \cite{gattupalli_iui2017}. There is a need for computational methods that would help with automatic computation of various physical performance metrics, so as to improve on the accuracy and efficiency of human-made assessments. Articulated hand pose recognition is an important step in recognizing and assessing physical exercises that contain hand gestures.

We discuss the selected hand gestures in Section \ref{rhythmic_hand_movements}, where we describe the physical exercise tasks and the importance of articulated hand pose estimation in assessing performance in those tasks. Recognition of rhythmic movements for rapid sequential hand gestures poses the additional challenges that the motion is fast and complicated. Moreover, the hand can be in any orientation and the dexterity of hand makes it hard to estimate and track finger positions.

The paper is further organized as follows: In Section 2, we discuss related work; Section 3 describes the rhythmic hand gestures we use in our dataset; in Section 4 we describe our HKD dataset; in Section 5, we discuss our experimental setup; in Section 6, we provide results and evaluations for some existing methods, and in Section 7 we provide conclusions and future work towards this problem.

\section{Related Work}
Most vision-based articulated hand pose estimation methods can be categorized based on the modality of the data that they use, their application towards first person view (ego-centric) or third person view, and whether the methods are discriminative or generative. Below we briefly provide a review of articulated hand pose estimation based on depth, color, or a combination of both depth and color modalities.

\subsection{Depth-based methods}
Hand pose estimation is frequently done using depth image frames. The use of depth cameras has facilitated the creation of large-scale datasets \citep{bighand2.2}. Automatic annotations of keypoint locations for such datasets can be achieved by attaching magnetic sensors to the hand. Using magnetic sensors for automated annotations is not as effective an approach for RGB datasets, as the magnetic sensors/ data gloves change the appearance of the hand in the RGB images and reduce the usefulness of such images for model training. This is why in this paper we focus on providing an RGB dataset. With the advent of deep learning based pose estimation, data has proven to be a crucial part of model learning. The model in \cite{5-DBLP:conf/iccvw/OberwegerL17} predicts 3D hand skeleton joint locations by integrating a principal component analysis (PCA) based prior into a Convolutional Neural Network (CNN). In \citet{2-conf/iccvw/KeskinKKA11}, 3D joint estimation is achieved by performing per pixel classification from synthetic depth images using random decision forests and has shown to achieve good recognition accuracy on ASL digits. Hierarchical pose regression for 3D hand poses is performed \cite{3-DBLP:conf/cvpr/0001WLT015} towards recognizing 17 ASL hand gestures. The work in \citet{7-DBLP:conf/cvpr/GeLYT16}, estimates 3D joint locations by projecting the depth image on 3 orthogonal planes and fusing the 2D regressions obtained on each plane to deduce the final estimate.  In \citet{8-7796640}, the hand segmentation task is performed using depth information to train a random decision forest. Hand pose estimation for tasks of hand-object interaction is performed in \citet{16-DBLP:conf/cvpr/HuangMMK15}. Here, grasp taxonomies are recognized for egocentric views of the hand. 

\subsection{RGB-based methods}
The methods in \citet{4-DBLP:conf/eccv/RogezKSMR14} and \citet{11-DBLP:conf/icassp/BaydounBMMRR17} contribute towards egocentric hand pose recognition from RGB image frames. Here, \citet{4-DBLP:conf/eccv/RogezKSMR14} is based on hierarchical regressors to estimate articulated hand pose and \citet{11-DBLP:conf/icassp/BaydounBMMRR17} formulates a graph based representation for hand shapes for hand gesture recognition. In \citet{10-DBLP:journals/pr/ZhouJL16} an RGB dataset of hands performing 10 different gestures is provided. The method in \citet{12-DBLP:conf/eccv/0002MZCOT16} consists of a model for color-based real-time hand segmentation and tracking for hands interacting with objects. In our work, we evaluate two state of the art hand keypoint detectors for 2D keypoint estimation from RGB images. The first one is the method presented in \citet{15-zb2017hand}, which proposes a deep neural network to perform 3D hand pose estimation. This model is trained on static images from a synthetic dataset. We evaluate the 2D keypoint detections from their PoseNet network on our dataset. The second state of the art hand keypoint detector that we evaluated is the method in \citep{openpose-simon2017hand}, which produces real-time keypoint detection on RGB images using multi-view bootstrapping. We define these methods in detail in Section 5 for our experiments and provide extensive evaluation for them in Section 6.

\subsection{RGB-D based methods}
The model in \citet{17-DBLP:journals/ijcv/TzionasBSAPG16} is a combined generative model with discriminatively trained salient features for articulated hand motion capture from monocular RGB-D images for hand-hand and hand-object interactions. The work in \citet{14-DBLP:journals/corr/Garcia-Hernando17} provides a new RGB-D dataset for egocentric hands manipulating with objects.  Another RGB-D approach for articulated 3D motion hand tracking is presented in \citet{13-DBLP:conf/iccv/SridharOT13} where the authors perform part-based pose retrieval and image-based pose optimization.

\section{Rapid Sequential Movement}
\label{rhythmic_hand_movements}
\begin{figure}[h]
 \centerline{\includegraphics[height=1.25in, keepaspectratio]{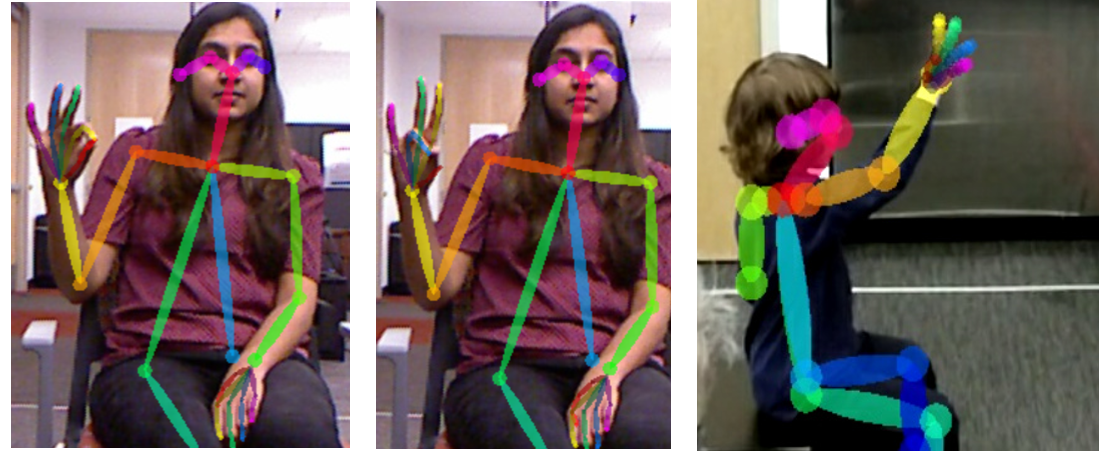}}
  \caption{Finger Appose hand and body keypoint detections by Method2 \cite{openpose-simon2017hand}}
  \label{fig:FingerTap}
\end{figure}
We present the Hand Keypoints Dataset (HKD) with images from participants performing gestures corresponding to ``finger appose" and ``appose finger succession" exercises.  These physical exercises are designed for cognitive behavior monitoring based on performance of rapid sequential movements. These exercises can be used as measures for fine motor skill development in children, and can be combined with tasks involving whole body motion for cognitive assessments \cite{gattupalli_iui2017}. These exercises consist of movements with hands and fingers and require wrist and finger keypoint detections. Here is a brief description of the exercises:

\subsection{Finger Tapping}
The participant is instructed to tap the index finger against the thumb as fast as possible. Recognition of these movements depends on accuracy of thumb and index fingertip keypoint ($TT and IT$ respectively) detections. If the Euclidean distance between $TT$ and $IT$ is less than a certain threshold we classify the frame as containing a tap movement.
\subsection{Appose Finger Succession}

\begin{wrapfigure}{l}{0.14\textwidth} 
\begin{center}
 \centerline{\includegraphics[height=0.65in, keepaspectratio]{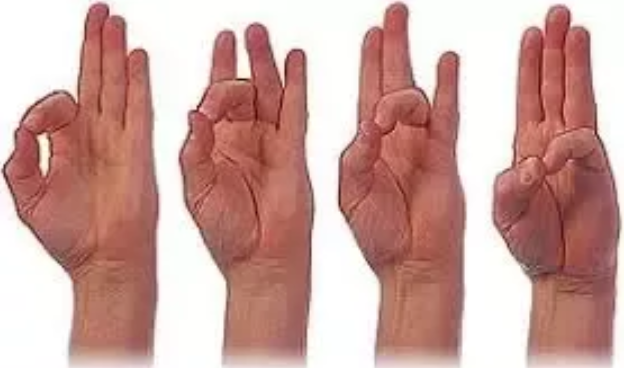}}
  \label{fig:appose}
  \caption{Appose Finger Succession}
  \vspace{-15pt}
\end{center}
\end{wrapfigure}
In this exercise, the participant is instructed to sequentially tap the index, middle, ring and little fingers (in that order) against the thumb. This sequence is to be repeated as as fast as possible. The participants are told not to tap in the backwards order. The keypoints features of most value here are the wrist and the five fingertips: Wrist(W), Thumb (TT), Index Finger tip (IT), Middle Finger tip (TM), Ring Finger tip (TR) and Little Finger tip (TL). The participant performance is evaluated based on speed and accuracy. 

Figure \ref{fig:FingerTap}, shows visualizations of the finger keypoints on participants performing finger tapping gestures. Figure 2, shows the appose finger successions that the participant is instructed to perform. The participant is told to freely change hand orientations with respect to camera while performing these movements. The distances between the detected hand keypoint locations can be used to understand the finger movements performed by the participant. In this work, we evaluate accuracy of these hand keypoint detections from single RGB image frames.

\section{Dataset}
We introduce the Hand Keypoints Dataset (HKD), which comprises annotated RGB images captured from participants while they perform rhythmic finger movements. Our dataset comprises of 782 color image frames captured from four different participants.
\begin{figure}[h]
    \vspace{-10pt}
 \centerline{\includegraphics[height=1.2in, keepaspectratio]{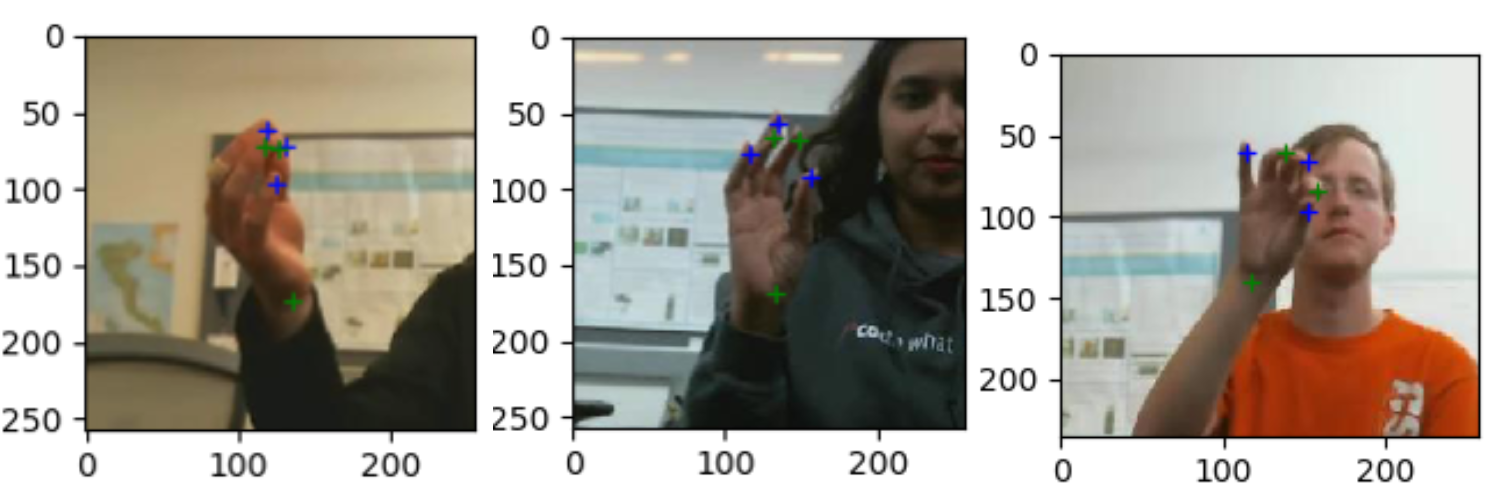}}
  \caption{Example annotations of cropped images from HKD dataset}
  \label{fig:hkd}
      \vspace{-10pt}
\end{figure}
The rapid sequential rhythmic finger movements are performed as described in Section \ref{rhythmic_hand_movements}. This is a novel benchmark dataset for hand keypoint detection and/or tracking from RGB images. We provide original frames with annotations as well as annotated cropped frames that are loosely cropped around the centroid location of the hand in the frame. We provide annotations for 6 hand keypoints namely - $W, TT, IT, TM, TR, TL$, for wrist and the tips of thumb, index, middle, ring and little finger respectively. We also provide the hand centroid location in the original RGB frames. For data collection, participants were told to perform the rapid sequential finger gestures specified in Section 3 and each participant performed these movements thrice with different hand orientations from the camera. The dataset consists of hand movements from four participants (2M, 2F) and the annotations are done manually by two annotators who used the annotation toolkit that we developed to perform standardized annotations.
\begin{figure}[h]
 \centerline{\includegraphics[height=2.6in, keepaspectratio]{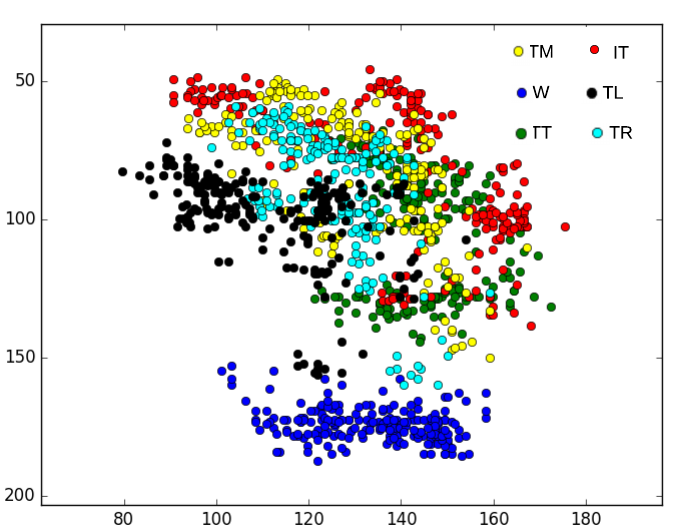}}
  \caption{Variance of hand keypoints of Subject1 in HKD.}
  \label{fig:variance}
    \vspace{-10pt}
\end{figure}
Additional details of our dataset, dataset images, annotations, annotation toolkit and code to display the annotations are available from:\\
\url{http://vlm1.uta.edu/~srujana/HandPoseDataset/HK_Dataset.html}. 
\begin{figure*}[!t]
\centering
  \includegraphics[height=3.8in,  keepaspectratio]{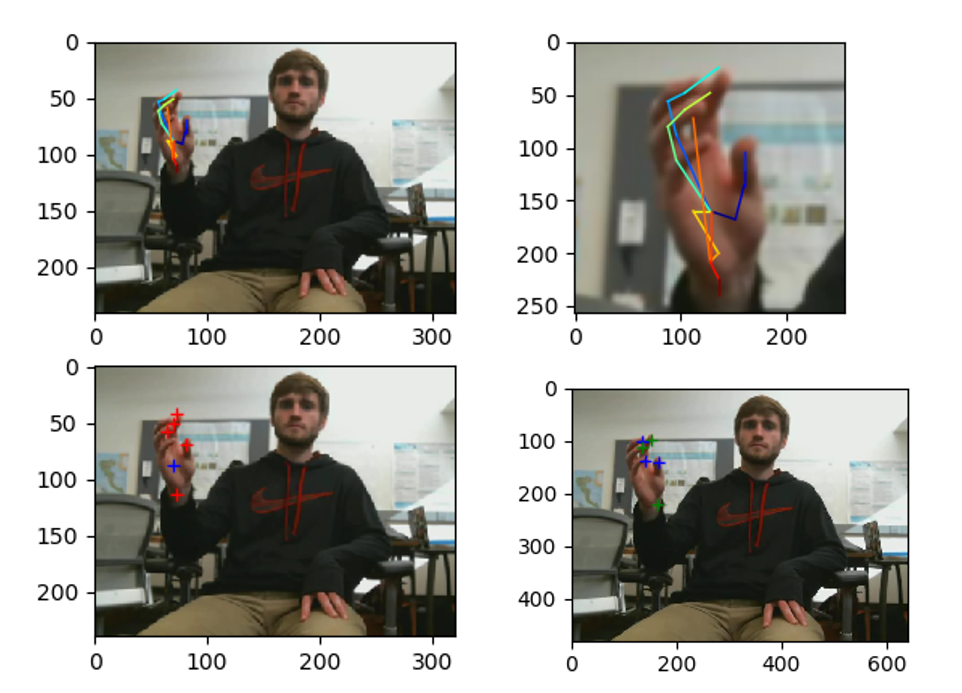}
 \caption{Visualization of Method1 \cite{15-zb2017hand} on HKD - Top Left, Top Right, Bottom Left. Bottom Right - Visualization of Ground Truth Annotations 6 keypoints.}
  \label{fig:view_method1}
\end{figure*}

\section{Experiments}
In our experiments, we evaluate the performance of deep learning based hand keypoint detectors on the HKD dataset. We perform experiments for current state of the art methods for hand keypoint localization for all 782 original images frames. We also provide subject-wise keypoint detection accuracy and per-keypoint accuracy for these methods, as described in Section \ref{results}.
\begin{figure}[h]
 \centerline{\includegraphics[height=1.4in, keepaspectratio]{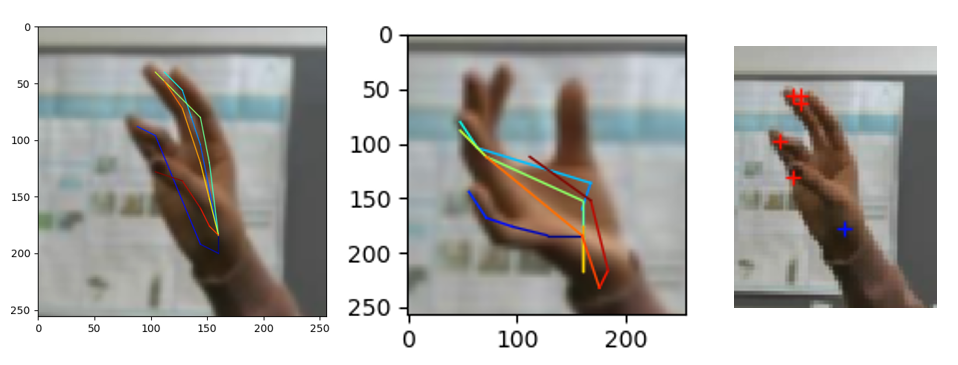}}
  \caption{Visualization of 21 keypoints estimated on HKD from Method1 \cite{15-zb2017hand} (left and center). Visualization of 6 keypoints of interests for our dataset.}
  \label{fig:method1}
\end{figure}

The first method (referred to as ``Method1''" in the experiments) that we have chosen for evaluation is the network in \citet{15-zb2017hand}. In this work, there is a hand segmentation network which is used to detect the hand and crop it. After this segmentation the image is passed as an input to the PoseNet network, which localizes 21 hand keypoints. We use the pre-trained model for hand segmentation network and the PoseNet provided with this work. This model is trained on the Rendered Handpose Dataset (RHD) which consists of 41258 synthetic RGB training images. We evaluate the performance of this network on original image frames from HKD for localization of our 6 keypoint of interest. The PoseNet network from this work is based on the work in \citet{cpm-conf/cvpr/WeiRKS16} and the output of the network consists of heatmaps (one heatmap for each keypoint).

The second method (referred to as ``Method2'') that we have chosen for evaluation is the network from \citet{openpose-simon2017hand}, which achieves state of the art performance on hand keypoint localization. The keypoint detection model architecture in this work is also similar to the work in \citet{cpm-conf/cvpr/WeiRKS16}. This is why we chose this method to be compared against method1 \cite{15-zb2017hand}. The difference here is that, for feature extraction, convolutional layers from a pre-initialized VGG-19 network are used as initial convolutional stages. The pretrained model from this paper is trained on hand data from multiple cameras using bootstrapping. This helps to correctly train the keypoints despite the occlusions from other hand joints as the same pose is observed from multiple cameras. This method also detects 21 keypoints, out of which we select the 6 keypoints annotated on our dataset. The pre-trained model from this work is trained on RGB images with multiple views (51 views) simultaneously at each timestamp. 2D keypoint detections from one frame are triangulated geometrically and used as training labels for a different view frame where that keypoint may be occluded. This is a boosted approach that learns keypoints labels from multiple views and improves on the supervised weak learners.

\subsection{Evaluation Protocol}
We provide metrics called average End Point Errors (EPE) in pixel for each keypoint and Percentage of Correct Keypoints (PCK) to measure accuracy. For each prediction, if the EPE is lower than a threshold then the keypoint is classified as correctly predicted. We evaluate the hand keypoint detection EPE on over several thresholds. We have set an upper limit to this threshold based on the detected keypoints. A reasonable upper limit to this threshold is measured as half of the average distance of each keypoint from centroid of the region of the hand in the image. Overall, EPE is shown in terms of Percentage of Correct Keypoints (PCK) and we plot the results on HKD dataset in Section 6. The hand region is determined by the convex hull formed by connecting the six annotated keypoints of the hand. Due to the large number of DOFs for the hand, we perform the following steps to find the threshold:
\begin{figure}[h]
    \vspace{-10pt}
 \centerline{\includegraphics[height=1.6in, keepaspectratio]{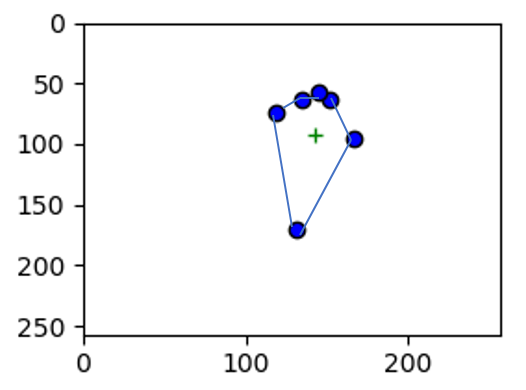}}
  \caption{Example visualization of convex hull from 6 keypoints. Green + is the centroid.}
  \label{fig:convex_hull}
    \vspace{-10pt}
\end{figure}
\begin{itemize}
\item For each image frame $i$, find the 2D convex hull from the six predicted keypoints. The convex hull of the set of keypoints would be the smallest convex polygon that contains all the six keypoints. Figure \ref{fig:convex_hull} shows a convex hull visualization.
\item Estimate centroid coordinate $C$, of the convex hull.
\item For each of the six precticed keypoints in $W, TT, IT, TM,$$ TR, TL$, find $d_w$, $d_{tt}$, $d_{it}$, $d_{tm}$, $d_{tr}$ and $d_{tl}$ as the Euclidean distances from $C$.

\item Let $d_i$ be average distance to centroid for that image frame, \[d_i = (d_w+ d_{tt}+ d_{it}+ d_{tm}+ d_{tr} +d_{tl})/6\].
\item Then, for $n$ being the number of frames (in our case $n=782$), threshold $T$ is computed as:
\[T = \frac{\sum_{i=1}^n d_i}{2\times n}\]
In this way, we employ a convex hull based approach to estimate an upper limit pixel threshold for evaluating correct keypoint detections. This value $T$ is 45  pixels for the HKD cropped dataset and 12 pixels for HKD original frames dataset. 
\end{itemize}
\section{Results}
\label{results}
\begin{figure}[h]
 \centerline{\includegraphics[height=2.2in, keepaspectratio]{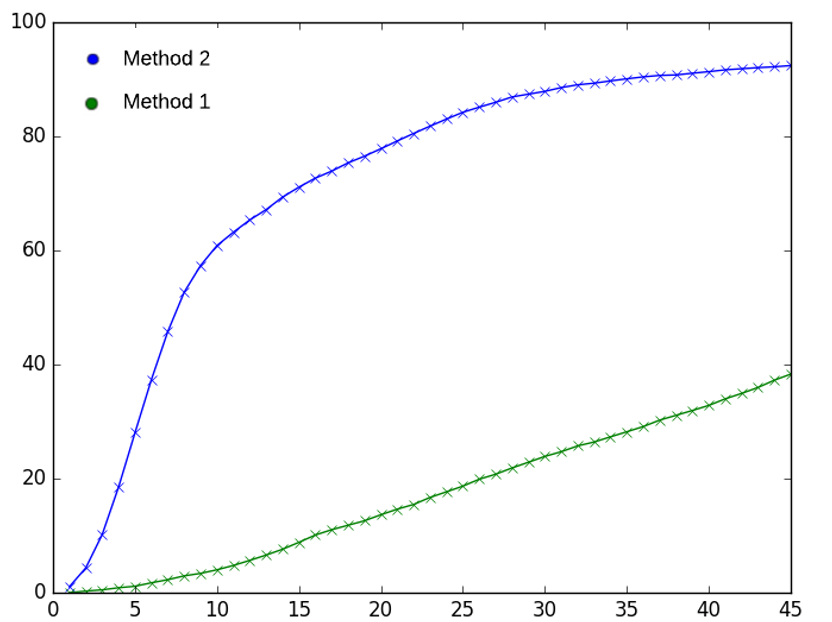}}
  \caption{Accuracy on HKD by Method1 (\citet{15-zb2017hand}) and 2 (\citet{openpose-simon2017hand}).}
  \label{fig:all_acc}
      \vspace{-5pt}
\end{figure}

We provide graphs to demonstrate the evaluation of Method1 \citet{15-zb2017hand} \& Method2 \citet{openpose-simon2017hand} on our dataset. Overall accuracy of both methods on HKD is shown in Figure \ref{fig:all_acc}. This plot demonstrates Percentage of Correct Keypoints (PCK) detections (y-axis) against different values of pixel thresholds (x-axis). For a threshold value of 11, the performance of the model from \citet{15-zb2017hand} is approximately 5\% and for the model from \citet{openpose-simon2017hand} is 60\%. 

\begin{figure}[h]
 \centerline{\includegraphics[height=2.2in, keepaspectratio]{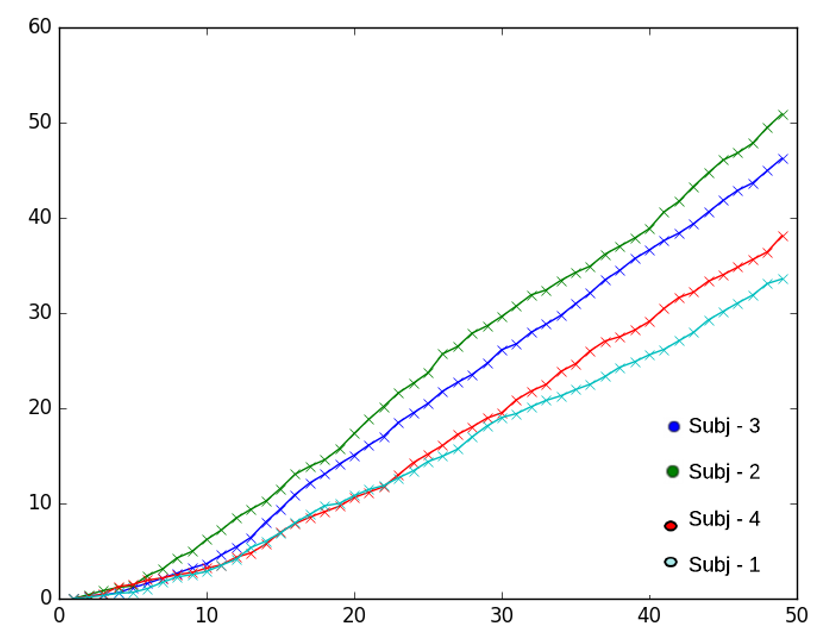}}
  \caption{Accuracy for each participant in HKD by Method1 (\citet{15-zb2017hand}).}
  \label{fig:subj_acc_method1}
      \vspace{-5pt}
\end{figure}

\begin{figure}[h]
 \centerline{\includegraphics[height=2.2in, keepaspectratio]{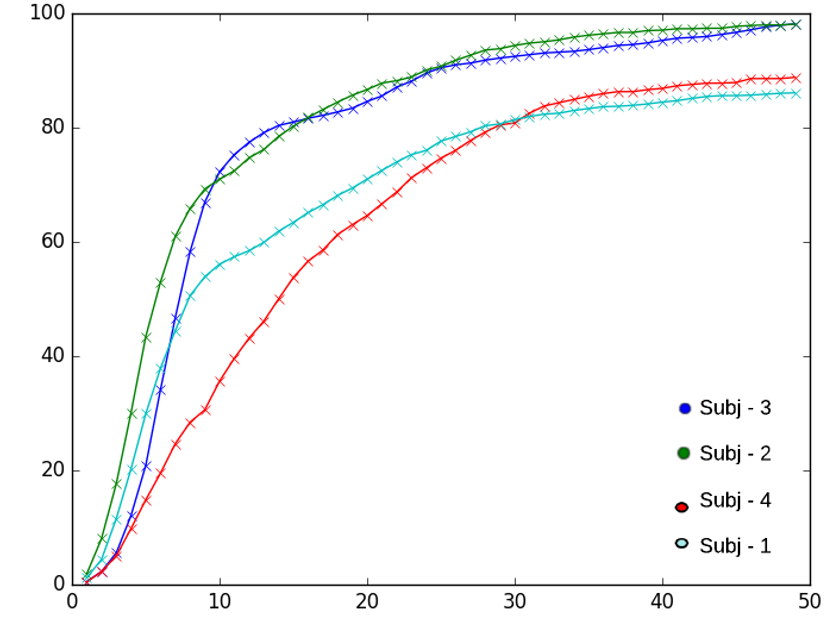}}
  \caption{Accuracy for each participant in HKD by Method2 (\citet{openpose-simon2017hand}).}
      \vspace{-5pt}
  \label{fig:subj_acc_method2}
\end{figure}

We provide accuracy attained on each participant's frame for Method1 \cite{15-zb2017hand} in Figure \ref{fig:subj_acc_method1}. Here, Probability (Percentage) of Correct Keypoints (PCK) vs. pixel thresholds is shown for each of four participants' frames of the HKD dataset for Method1. Similarly, in Figure \ref{fig:subj_acc_method2} we plot PCK vs. pixel thresholds to show accuracy for each of the four HKD participants' frames separately for Method2 \citet{openpose-simon2017hand}. These are important benchmarks for performing subject independent tests on HKD. The results demonstrated by these plots shows that the performance for both these methods on subject 3 is higher than for other participants. The performance for subjects 3 and 4 is approximately 10-15\% lower than for other subjects. 

\begin{figure}[h]
 \centerline{\includegraphics[height=2.2in, keepaspectratio]{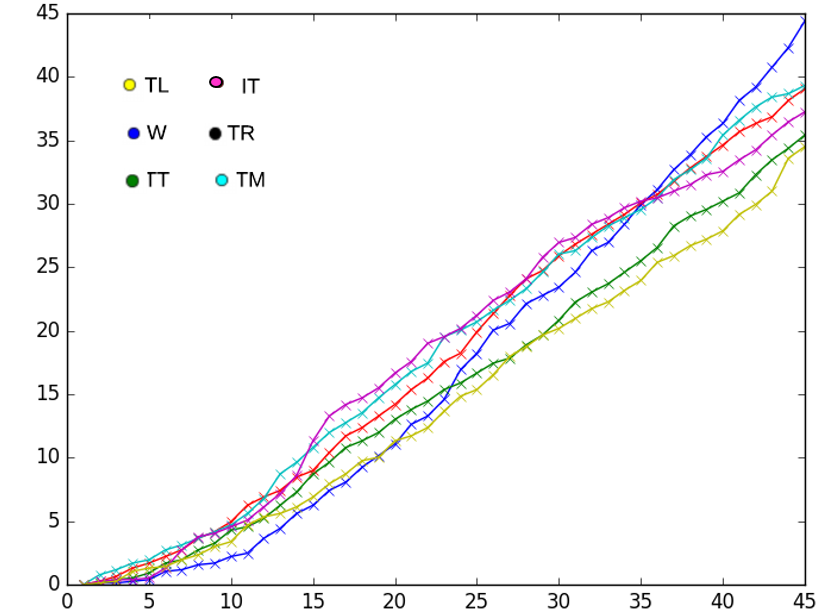}}
  \caption{Accuracy for each of 6 keypoints by Method1 (\citet{15-zb2017hand}).}
  \label{fig:IInd_acc_method1}
    \vspace{-5pt}
\end{figure}

\begin{figure}[h]
 \centerline{\includegraphics[height=2.2in, keepaspectratio]{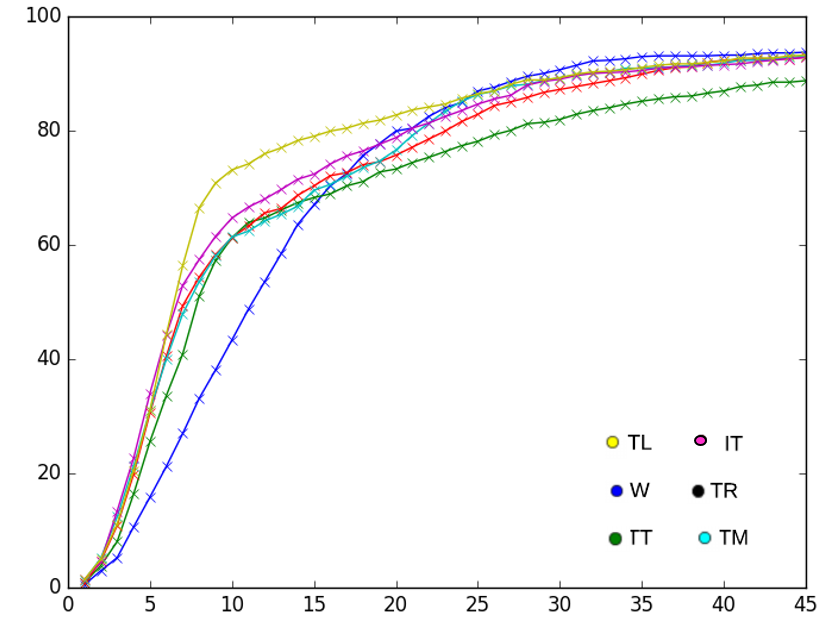}}
  \caption{Accuracy for each of 6 keypoints by Method2 (\citet{openpose-simon2017hand}).}
  \label{fig:IInd_acc_method2}
    \vspace{-10pt}
\end{figure}

We also measure accuracy attained for each of the six keypoints ($W, TT, IT, TM, TR, TL$) individually in Figure \ref{fig:IInd_acc_method1} and \ref{fig:IInd_acc_method2}. The plot in Figure \ref{fig:IInd_acc_method1} demonstrates per keypoint PCK vs pixel threshold for method1 \cite{15-zb2017hand} and the plot in Figure \ref{fig:IInd_acc_method2} shows per keypoint PCK vs pixel threshold for method2 \cite{openpose-simon2017hand} performance. These plots help contrast the performances for both methods for each of the six keypoints. For both methods, wrist detection performance is better than detection for finger tip keypoints. Overall, the method in paper \citet{openpose-simon2017hand} outperforms the method in paper \citet{15-zb2017hand} but the trends in PCK for subject-wise and keypoint-wise accuracies are similar. The method in paper \citet{openpose-simon2017hand} leverages the recognition by a boosted approach by improving accuracy of trained classifier by refinement of weakly supervised learning using multiview geometry. Here, occluded keypoints in one view frames are projected from another camera view frame where they are visible and the training is refined with the help of these new labels. The pre-trained model for this method has been trained on these multiview image frames of RGB hand images. In contrast, the hand keypoint detector in paper \citet{15-zb2017hand} is trained on RGB synthetic images from Rendered Handpose Dataset (RHD). The results can also be an indicator that higher performance accuracy can be achieved by training a model on real RGB images performance as compared to training on images from a synthetic dataset. 

\section{Conclusions and future work}
Vision-based hand keypoint detection is important for several applications and it is a challenge to solve the problem without depth cameras, color gloves or other sensors. As a part of this work, we have presented a hand keypoint recognition benchmark, HKD, that would serve as a baseline towards keypoint detection for rapid sequential movements. In this paper, we have discussed two current state of the art methods and have provided quantitative evaluations of these methods on our hand keypoint detection dataset. We have designed our dataset to encompass rapid sequential movements from gesture-based exercises used for cognitive behavior assessments. We briefly discussed these hand and finger movements in Section \ref{rhythmic_hand_movements}. As a part of future work, we plan to work towards automated algorithms that can analyze and score participant performances in such exercises, based on deep visual features. One important aspect of this is to build algorithms to dynamically classify the hand postures using the keypoints.
\begin{figure*}[!t]
\centering
  \includegraphics[height=3.5in,  keepaspectratio]{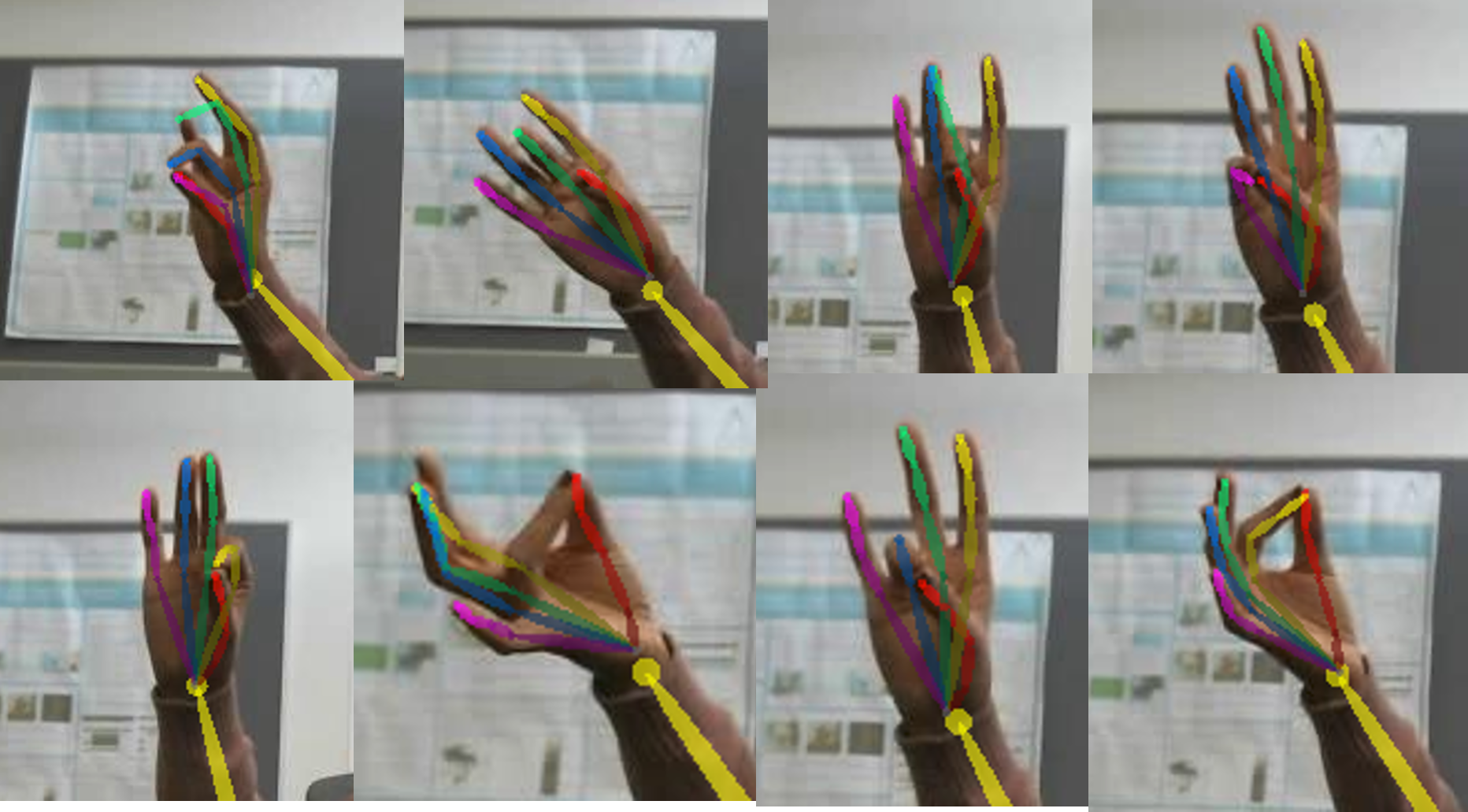}
 \caption{Visualization of Method2 (model from \citet{openpose-simon2017hand}) on HKD.}
  \label{fig:view_method2}
\end{figure*}

\begin{acks}
This work is partially supported by National Science Foundation grants CNS-1338118, IIS-1565328 and IIP 1719031. Any opinions, findings, and conclusions or recommendations expressed in this publication are those of the authors, and do not necessarily reflect the views of the National Science Foundation.
\end{acks}

\bibliography{sigchi.bib}

\bibliographystyle{ACM-Reference-Format}

\end{document}